\def\year{2019}\relax
\pdfoutput=1
\documentclass[letterpaper]{article} %DO NOT CHANGE THIS
\usepackage{aaai19}  %Required
\usepackage{times}  %Required
\usepackage{helvet}  %Required
\usepackage{courier}  %Required
\usepackage{url}  %Required
\usepackage{graphicx}  %Required
\usepackage{amssymb}
\usepackage{amsmath}
\usepackage{algorithm}
\usepackage[noend]{algpseudocode}
\usepackage{enumitem}
\usepackage{graphicx}
\usepackage{subfig}
\usepackage{comment}
\begin{document}
\title{End-to-End Refinement Guided by Pre-trained Prototypical Classifier}
\author{
Junwen Bai$^1$,
Zihang Lai$^2$,
Runzhe Yang$^3$, 
Yexiang Xue$^4$, 
John Gregoire$^5$,
Carla Gomes$^1$
\\ 
$^1$ Cornell University ~~~ $^2$ Oxford University ~~~ $^3$ Princeton University \\
$^4$ Purdue University ~~~ $^5$ California Institute of Technology \\
\{jb2467,zl723,ry289,yx247\}@cornell.edu,
gregoire@caltech.edu,
gomes@cs.cornell.edu,
}
\maketitle
\begin{abstract}
Many real-world tasks involve identifying patterns from data satisfying background or prior knowledge. In domains like materials discovery, due to the flaws and biases in raw experimental data, the identification of X-ray diffraction patterns (XRD) often requires a huge amount of manual work in finding refined phases that are similar to the \textit{ideal} theoretical ones. Automatically refining the raw XRDs utilizing the simulated theoretical data is thus desirable. We propose \textit{imitation refinement},   a novel approach  to refine  imperfect input patterns, guided by a pre-trained classifier incorporating prior knowledge from simulated theoretical data, such that the refined patterns \textit{imitate} the \textit{ideal} data. The \textit{classifier} is trained on the \textit{ideal} simulated data to classify patterns and learns an embedding space where each class is represented by a \textit{prototype}. The  \textit{refiner} learns to refine the imperfect patterns with small modifications, such that their embeddings are closer to the corresponding prototypes. We show that the refiner can be trained in both supervised and unsupervised fashions. We further illustrate the effectiveness of the proposed approach both qualitatively and quantitatively in a digit refinement task and an X-ray diffraction pattern refinement task in materials discovery. 
\end{abstract}

\section{Introduction}

Many real-world tasks involve identifying meaningful patterns satisfying background or prior knowledge  from limited amount of labeled data \cite{chapelle2009semi}. Furthermore, the raw data are often corrupted with noise \cite{steinbrener2010data}, which makes it even harder to identify meaningful patterns. On the other hand, in many domains like scientific discovery, though the experimental data might be flawed or biased, \textit{ideal} data can often be synthesized easily \cite{rubin1993discussion,bernstein2014computational}.  It is thus desirable to incorporate knowledge from  \textit{ideal} data to refine the quality of the raw patterns to make them more meaningful and recognizable.

\begin{figure}[t]
	\centering
	\includegraphics[width=0.95\linewidth]{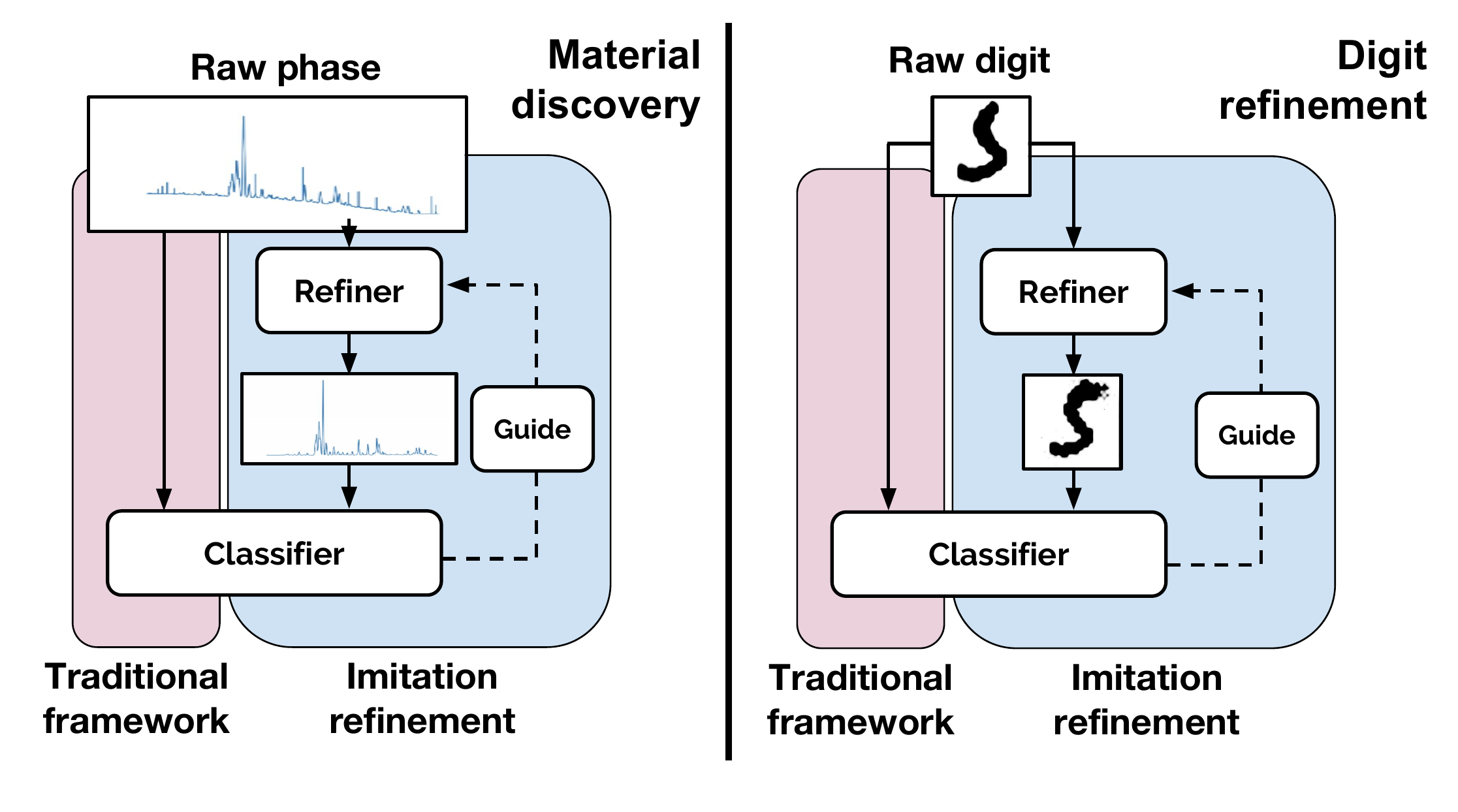}
    \vspace{-1em}
    \caption{Imitation refinement improves the quality of imperfect patterns guided by a pre-trained classifier incorporating prior knowledge from ideal patterns. Left: Refinement of XRD patterns. Right: Refinement of hand-written digits. }
	\label{fig:intro}
    \vspace{-1em}
\end{figure}

For instance, in materials discovery, where we would like to discover new materials, each material is characterized by a unique X-ray diffraction pattern (also called XRD or phase, see Fig. \ref{fig:intro}). 
%The identification of such phases is challenging because not only the raw phases from experiments are often mixed with others and further corrupted by noise, but also raw phases data are provided with very little supervision (e.g. crystal structure group) or even no supervision. Furthermore, besides predicting the properties of the materials from the raw imperfect phases such as crystal structure group classification tasks \cite{park2017classification}, 
The identification of such phases is challenging because the raw phases from experiments are often mixed with each other and further corrupted with noise. Moreover, material scientists are interested in not only predicting the properties of materials \cite{park2017classification}, but also finding refined phases that are of better quality and similar to the ideal theoretical phases \cite{speakman2013introduction}. To the best of our knowledge, this task can only be computed manually using quantum mechanics, which often requires huge amount of manual work even for an expert. 

Even though our work has been motivated by applications in scientific discovery, there are other domains in which imitation refinement is applicable. For example, in the context of digit recognition,  some scratchy hand-written digits may be  hard to read since they may  miss important strokes or are poorly written. Given the labels of the digits and synthesized ``ideal" typeset digits, one may want  to refine the hand-written digits to improve their readability, though typically we do not know what the corresponding ground-truth ideal digits are.
%Classical end-to-end models \cite{xie2012image} cannot be applied here since the raw and ground-truth pattern pairs are not provided. Instead, the task is to refine an imperfect pattern given its class label if available, or if the class label is not available, the imperfect pattern should be refined according to the most likely class label. 

%An efficient method may be needed to automatically recognize and refine imperfect patterns.
We propose a novel approach called \textbf{imitation refinement}, which improves the quality of imperfect patterns by \textbf{imitating} ideal patterns, guided by a classifier with prior knowledge pre-trained on the ideal dataset.
Imitation refinement applies small modifications to the imperfect patterns such that (1) the refined patterns have better quality and are similar to the ideal patterns and (2) the pre-trained classifier can achieve better classification accuracy on the refined patterns. We show that both ends can be achieved even with limited amount of data.
% With only limited amount of imperfect data, we show that both ends can be achieved. 

Specifically, we pre-train a classifier using the ideal synthetic patterns and learn a meaningful embedding space. In such a space, each class forms a cluster containing all the embedded inputs from this class. We call the cluster centers \textit{prototypes} for each class. Then the refiner learns a mapping from imperfect patterns to refined patterns, such that the embeddings of the refined patterns are closer to the corresponding prototypes and give better prediction results. In the supervised case, the corresponding prototype is the one associated with the class. In the unsupervised case, the prototype is the closest one to the raw embedded input.

The main contribution of our work is to provide a novel framework for imitation refinement, which can be used to improve the quality of imperfect patterns under the supervision from a classifier containing prior knowledge. Our second contribution is to find an effective way to incorporate the prior knowledge from the ideal data into the classifier. The third contribution of this work is to provide a way to train the refiner even if the imperfect inputs have no supervision. 

Using a materials discovery dataset, we show that for the imperfect input experimental phases, the refined phases are closer to the quantum-mechanically computed ideal phases. In addition, we achieve higher classification accuracy on the refined phases. We show that even in the unsupervised case, the refinement can help improve the quality of the input patterns. To validate the generality of our approach, we also show that  imitation refinement improves the quality of poorly written digits by imitating ideal typeset digits.

\section{Imitation Refinement}

\subsection{Notation}
In imitation refinement, we are given an ideal dataset $\mathcal D_{ideal}=\{(x_i^{ideal}, y_i^{ideal})\}_{i=1}^N$. The ideal $d$-dimensional features $x_i^{ideal}\in \mathcal X^{ideal} \subseteq \mathbb R^d$ is a realization from a random variable $X^{ideal}$, and the label $y^{ideal}_i \in \mathcal Y$, where $\mathcal Y$ is a discrete set of classes $\{0,1,...,l\}$ in this problem. In addition, we are also given the imperfect training data $\mathcal D_{imp}$. In the supervised/targeted case, $\mathcal D_{imp}=\{(x^{imp}_i, y^{imp}_i)\}_{i=1}^{M}$ where $x_i^{imp}\in \mathcal X^{imp} \subseteq \mathbb R^d$ is a realization of a random variable $X^{imp}$ and the labels $y^{imp}_i\in \mathcal Y$. In the unsupervised/non-targeted case, $\mathcal D_{imp}=\{x^{imp}_i\}_{i=1}^{M}$ where $x^{imp}\in \mathcal X^{imp}$ and the labels are not available. We assume there is a slight difference \cite{shimodaira2000improving,sugiyama2012machine} between $\mathcal X^{imp}$ and $\mathcal X^{ideal}$. 

\subsection{Problem Description}

Our goal is  to learn a function $\mathcal R: \mathcal X^{imp} \to \mathcal X^{rfd}$,
where $\mathcal X^{rfd} \subseteq \mathcal X^{ideal},$ 
%such 
that 
%this function 
refines the imperfect patterns 
%to resemble 
into ideal patterns 
%in $\mathcal X^{ideal}$ 
(e.g. the theoretically computed corresponding patterns), with the guidance from a pre-trained classifier $\mathcal C$. $\mathcal C$ is the composition $\mathcal G_{\psi} \circ \mathcal F_{\theta}$ where $\mathcal F_{\theta}: \mathcal X^{ideal} \to \mathbb R^m$ is an embedding function ($m$-dimensional embedding space) and $\mathcal G_{\psi}: \mathbb R^m \to \mathcal Y$ is a prediction function. For the inputs $x$, we hope $\mathcal C(\mathcal R(x))$ can give better results than $\mathcal C(x)$, and $\mathcal R(x)$ has better quality than $x$, by imitating the patterns in $\mathcal X^{ideal}$.

%We first train a classifier on the ideal dataset to incorporate prior knowledge. More specifically, we desire to learn an embedding function $\mathcal F_{\theta}: \mathcal X^{ideal} \to \mathbb R^m$ and a prediction function $\mathcal G_{\psi}: \mathbb R^m \to \mathcal Y$. Then the classifier $\mathcal C: \mathcal X^{ideal} \to \mathcal Y$ is thus the combination $\mathcal G_{\psi} \circ \mathcal F_{\theta}$. Function $\mathcal G_{\psi}$ could be the sigmoid function, softmax function, the last softmax layer of neural networks, etc..

\subsection{Pre-trained Prototypical Classifier}

\begin{figure}[t]
	\centering
	\includegraphics[width=0.5\textwidth]{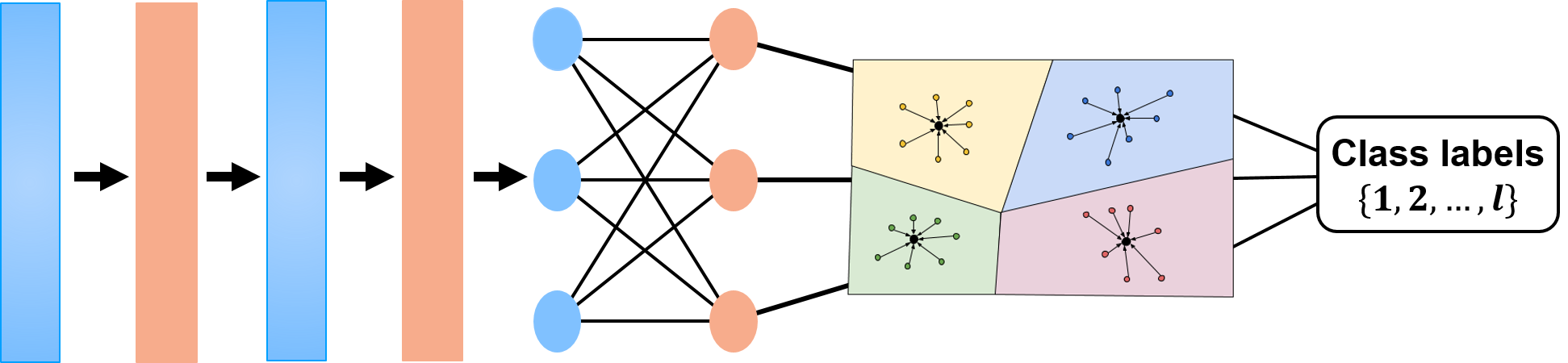}
    \caption{Prototypical classifier not only predicts labels, but also learn a meaningful embedding space. The center of the cluster for each class is called prototype.}
	\label{fig:cls}
    \vspace{-1em}
\end{figure}

Inspired by recently proposed prototypical networks \cite{snell2017prototypical}, the classifier is trained to learn a meaningful embedding space to better incorporate the prior knowledge as well as the class prediction. The embedding space is formed by the features from the last layer before the softmax layer, where each class can be represented by a prototype embedding and embeddings from each class form a cluster surrounding the prototype. We thus call the classifier \textit{prototypical classifier}. The ideal dataset is used to train the prototypical classifier and inject prior knowledge into the classifier.  The prototype of each class is the mean of the embedded patterns from this class:
\begin{equation} \label{ck}
c_k = \frac{1}{|\mathcal D^{ideal}_k|}\sum_{y_i^{ideal}=k} \mathcal F_{\theta}(x_i^{ideal})
\end{equation}
where $c_k$ is the prototype of class $k$ and $\mathcal D^{ideal}_k$ is the subset of $\mathcal D_{ideal}$ containing all the samples from class $k$. 

To learn such a prototypical classifier, besides the class prediction loss given by a classification loss $\mathcal L_{\mathcal C}(\theta, \psi)=\sum_i \ell(\mathcal G_\psi(\mathcal F_\theta(x^{ideal}_i)), y_i^{ideal})$ where $\ell(\cdot)$ can be the cross entropy loss or other supervised losses, we further add a loss defining the distances to the ground-truth prototypes in the embedding space given a distance function $d: \mathbb R^m \times \mathbb R^m \to [0, +\infty)$:
\begin{equation}
\mathcal L_{\mathcal F}(\theta)=\sum_i -\log \frac{\exp(-d(\mathcal F_\theta(x_i^{ideal}), c_{y_i^{ideal}}))}{\sum_{k'\in \mathcal Y}\exp (-d(\mathcal F_\theta(x_i^{ideal}), c_{k'}))}
\end{equation}
The idea behind this loss is simple: for a sample $x_i^{ideal}$, we define a distribution over classes based on a softmax over the distances to the prototypes and the loss is simply the negative log-likelihood of the probabilities. In each training step, the batch of samples are randomly selected from each class to ensure each class has a least one sample. The prototype of each class is randomly initialized before the training and is updated by computing the mean of the embeddings from the class and the prototype from last batch. During the training, we also consider the prototypes from the last batch while updating the prototypes to stabilize the prototypes instead of learning new prototypes in each step. If we assume the embedding space is well formed by clusters (which will be shown in experimental section), cluster means are the best representatives as shown in \cite{banerjee2005clustering}. Pseudocode to train the prototypical classifier is provided in Algorithm \ref{proto_cls}. Fig. \ref{fig:cls} gives an overview of the prototypical classifier.

\begin{algorithm}
\caption{One epoch in the training for the prototypical classifier}\label{proto_cls}
\textbf{Input:} Ideal dataset $\mathcal D_{ideal}=\{(x_i^{ideal}, y_i^{ideal})\}_{i=1}^N$ where $y_i^{ideal}\in \{1,...,l\}$, max number of batches($T$), the number of samples selected from each class in each step($N_c$), the subset $\mathcal D_k$ containing all the samples from class $k$.\\
\textbf{Output:} Prototypical classifier $\mathcal C_{\theta, \psi}$.
\begin{algorithmic}[1]
\State Randomly initialize prototypes $c^0_k$ for each class.
\For{$t=1,...,T$}
\For{$k=1,...,l$}
\State Randomly select $N_c$ samples from class $k$.
\EndFor
\State Compute $\mathcal L_{\mathcal C}(\theta, \psi)$ and $\mathcal L_{\mathcal F}(\theta)$ given the samples and prototypes.
\State $\mathcal L_{batch}(\theta, \psi)\gets \mathcal L_{\mathcal C}(\theta, \psi)+\lambda \mathcal L_{\mathcal F}(\theta)$.
\State Update the parameters $\theta, \psi$ by taking an Adam step on the batch loss.
\State $c_k^t \gets \frac{1}{|\mathcal D^{ideal}_k|+1}(c^{t-1}_k+\sum_{y_i^{ideal}=k} \mathcal F_{\theta}(x_i^{ideal}))$
\EndFor
\end{algorithmic}
\end{algorithm}

\subsection{Imitation Refiner}

\begin{figure*}[t]
	\centering
	\includegraphics[width=0.8\textwidth]{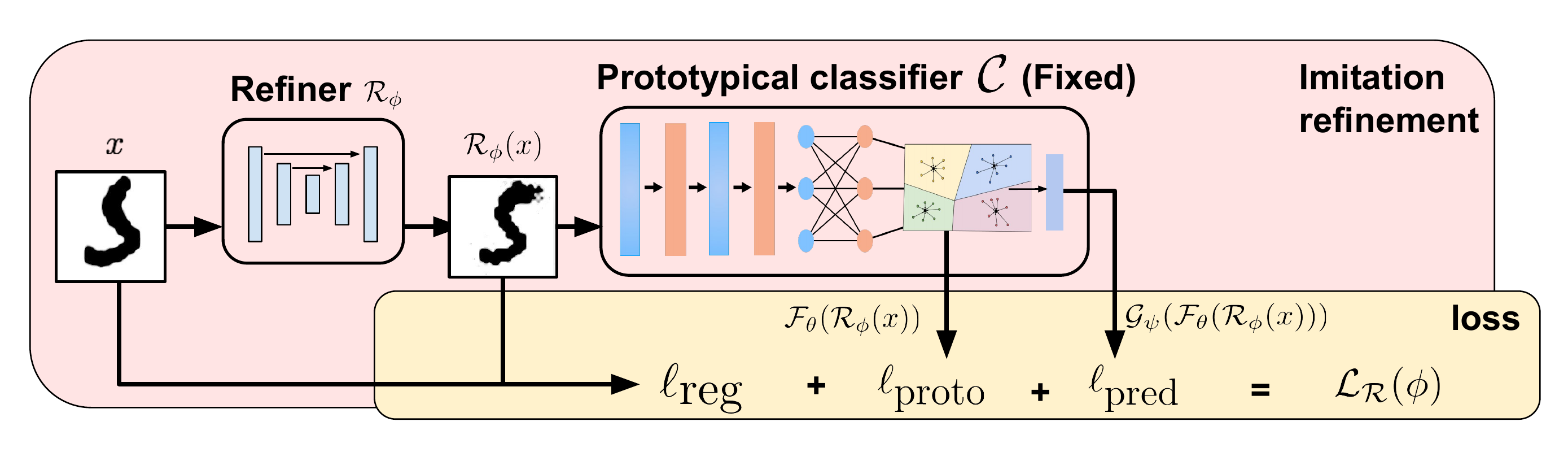}
    \vspace{-1em}
    \caption{The refiner $\mathcal R_\phi$ is trained in an end-to-end fashion. The Pre-trained classifier $\mathcal C$ provides two losses, $\ell_{\mbox{proto}}$ and $\ell_{\mbox{pred}}$. Loss $\ell_{\mbox{reg}}$ is given by the difference between the refined input and the raw input in either the raw space or some feature space.} 
	\label{fig:refiner}
    \vspace{-1em}
\end{figure*}

The pre-trained prototypical classifier is then applied to guide the training of the refiner $\mathcal R_\phi: \mathcal X^{imp} \to \mathcal X^{rfd}$ with learnable parameters $\phi$. We propose to learn $\phi$ by minimizing a combination of three losses:
\begin{equation}\label{rf_loss}
\begin{aligned}
\mathcal L_R(\phi) =& \sum_i (\ell_{\mbox{pred}}(\phi; x_i^{imp}, \mathcal Y) + \alpha \ell_{\mbox{reg}}(\phi;x_i^{imp})\\
&+ \beta \ell_{\mbox{proto}}(\phi; x_i^{imp}, \mathcal F(\mathcal X^{ideal}))).
\end{aligned}
\end{equation}
where $x_i^{imp}$ is the $i^{th}$ imperfect training sample and $\mathcal F(\mathcal X^{ideal})$ is the embedding space formed by the embeddings of samples from the space $\mathcal X^{ideal}$. $\alpha$ and $\beta$ are coefficients that trade off different losses. Note that once the classifier $\mathcal C$ is trained, it is fixed along with the prototypes $c_k$'s during the training of the refiner. Furthermore, $
\mathcal C$ provides loss functions ($\ell_{\mbox{pred}}$ and $\ell_{\mbox{proto}}$) to the training of $\mathcal R_\phi$.

As we mentioned previously, the refiner can be trained in both targeted and non-targeted fashions, depending on whether the labels of the imperfect training samples are provided or not.
In the targeted case, prediction loss $\ell_{\mbox{pred}}$ is the loss given by the difference between the predicted labels of the refined input patterns and the ground-truth labels:
\begin{equation}
\ell_{\mbox{pred}}(\phi;x_i^{imp},\mathcal Y)=\mathcal H(\mathcal C(\mathcal R_\phi(x_i^{imp})), y_i^{imp})
\end{equation}
where $\mathcal H$ is the cross-entropy loss. In the non-targeted case, we simply change the cross-entropy loss to the entropy loss, $\mathcal H(\mathcal C(\mathcal R_\phi(x_i^{imp})))$, to represent the uncertainty of the classifier $\mathcal C$ on the refined patterns. The goal is to minimize the entropy to force the refiner to learn more meaningful refined patterns which could be better recognized by $\mathcal C$.

Simply using the prediction loss is not sufficient to learn an effective refiner and sometime might learn adversarial examples \cite{kurakin2016adversarial}. We thus introduce the prototypical loss $\ell_{\mbox{proto}}$ to further guide the refinement towards the corresponding prototypes in the embedding space, which is more robust. In the targeted case, the prototypical loss is given by the negative log-likelihood  on the distances between the embeddings of the refined patterns and the ground-truth prototypes:
\begin{equation}
\begin{aligned}
&\ell_{\mbox{proto}}(\phi; x_i^{imp}, \mathcal F(\mathcal X^{ideal}))\\
=&-\log \frac{\exp(-d(\mathcal F(\mathcal R_\phi(x_i^{imp})), c_{y_i^{imp}}))}{\sum_{k'}\exp (-d(\mathcal F(\mathcal R_\phi(x_i^{imp})), c_{k'}))}
\end{aligned}
\end{equation}
In the non-targeted case, we use entropy loss:
\begin{equation}
\begin{aligned}
\ell_{\mbox{proto}}(\phi; x_i^{imp}, \mathcal F(\mathcal X^{ideal}))=\sum_{k=1}^l -p^{imp}_{i,k} \log{p^{imp}_{i,k}}
\end{aligned}
\end{equation}
where $p_{i,k}^{imp}= \frac{\exp(-d(\mathcal F(\mathcal R_\phi(x_i^{imp})), c_k))}{\sum_{k'}\exp (-d(\mathcal F(\mathcal R_\phi(x_i^{imp})), c_{k'}))}$.

\begin{algorithm}
\caption{One epoch in the training for the refiner when labels are available}\label{rf_tar}
\textbf{Input:} Imperfect training dataset $\mathcal D_{imp}=\{(x_i^{imp}, y_i^{imp})\}_{i=1}^M$ where $y_i^{imp}\in \{1,...,l\}$, max number of batches($T$), the batch size($N_c$), the prototypical classifier $\mathcal C$ and the prototypes $c_k$.\\
\textbf{Output:} Refiner $\mathcal R_\phi$.
\begin{algorithmic}[1]
\For{t=1..T}
\State Sample $N_c$ samples from training set $D_{imp}: \{x_i, y_i\}_{i=1}^{N_c}$.
\State Let $r_i=\mathcal R(x_i)$ be the refined inputs.
\State Let $e_i=\mathcal F(r_i)$ be the embedded refined inputs.
\State Let $c_i=\mathcal G(e_i)$ be the predicted labels.
\State Compute $\mathcal L_{\mathcal R}(\phi)$ in equation (\ref{rf_loss})
\begin{itemize}[leftmargin=.5in]
\item $\ell_{\mbox{pred}}=\frac{1}{N_c}\sum_i\mathcal H(c_i, y_i)$
\item $\ell_{\mbox{proto}}=\frac{1}{N_c}\sum_i -\log \frac{\exp(-d(e_i, c_{y_i}))}{\sum_{k'}\exp (-d(e_i, c_{k'}))}$
\item $\ell_{\mbox{reg}}=\frac{1}{N_c}\sum_i ||\Psi(r_i)-\Psi(x_i)||_p$
\end{itemize}
\State Update parameters $\phi$ through back-propagation based on the loss $\mathcal L_{\mathcal R}(\phi)$.
\EndFor
\end{algorithmic}
\end{algorithm}

Note that for an imperfect sample $x_i^{imp}$, we are looking for an ideal pattern in $\mathcal X^{ideal}$ that is most related to $x_i^{imp}$. $\mathcal R$ should modify the input as little as possible to remain the contents in the imperfect input samples \cite{gatys2016image}. Therefore, we introduce the third loss $\ell_{\mbox{reg}}$ to regularize the changes made for the input:
\begin{equation}
\ell_{\mbox{reg}}(\phi; x^{imp}_i)=||\Psi(\mathcal R(x_i^{imp}))-\Psi(x_i^{imp})||_p
\end{equation}
where $||\cdot||_p$ is p-norm and $\Psi$ maps the raw input into a feature space. $\Psi$ can be an identity map $\Psi(x)=x$ or more abstract features such as the feature maps after the first or second convolution layer. This loss works for both targeted and non-targeted cases since it does not rely on the labels.
Such regularization can also help avoid learning an ill-posed mapping from $\mathcal X^{imp} $ to $\mathcal X^{ideal}$ such as a many-to-one mapping that maps all the imperfect patterns from class $k$ to one ideal pattern in class $k$ regardless the raw contents in the imperfect patterns. This mapping could achieve very small $\ell_{\mbox{pred}}$ and $\ell_{\mbox{proto}}$ but that is not what we want.

The refiner $\mathcal R$ is trained in an end-to-end way as described above and all the parameters are updated through back-propagation. The pseudocode to train the refiner is provided in Algorithm \ref{rf_tar} in the targeted case. In the non-targeted case, the algorithm is simply replacing the targeted losses with non-targeted losses as described. The overall structure of the refiner is shown in Fig. \ref{fig:refiner}.

\section{Experiments}

In this section we present results to validate our approach on two applications: materials discovery \cite{jain2013commentary} and hand-written digits refinement \cite{mnist}.

\subsection{Materials Discovery}

High-throughput combinatorial materials discovery is a materials science task whose intent is to discover new materials using a variety of methods including X-ray diffraction pattern analysis\cite{green2017fulfilling}. The raw imperfect X-ray diffraction patterns (XRD) from experiments are often unsatisfiable because the data corruption could happen in any step of the data processing. Much effort has been put into cleaning the data through techniques like matrix decomposition, data smoothing \cite{chen2005enhancing,suram2016automated}. These previous works mainly focus on individual pattern cleaning instead of modifying the pattern by considering prior knowledge. Thus, experts still need to do considerable manual work to fit the data into the heavy-duty quantum mechanical computation to find a refined XRD similar to some perfect theoretical pattern, which may take weeks or even months. However, some domain knowledge can be very useful to automatically push the refinement of the imperfect XRD patterns. For example, it is a fundamental fact that each XRD could be categorized as exactly 1 of the 7 crystal structures (triclinic, monoclinic, orthorhombic, tetragonal, rhombohedral, hexagonal, cubic). Each structure has some unique signal patterns. We want to learn such knowledge from the ideally simulated data and further guide the refinement of the imperfect XRD patterns.

In this work, we show how close the refined XRDs are to the quantum-mechanically computed patterns to validate that useful domain knowledge is learned by the pre-trained classifier. We show our performance via two metrics. First, the refined XRDs can achieve better classification accuracy even if the classifier is not changed. Second, we directly show the improvement of the quality of the refined XRDs both qualitatively and quantitatively. We measure the difference between the ground-truth XRDs and refined XRDs on $\ell_1, \ell_2$, KL-divergence and cross correlation. Qualitative results are also shown in the heatmaps of the XRDs.

\noindent \textbf{Dataset:} The dataset used in this application is from materials project \cite{jain2013commentary}. The ideal simulated data have approximately 240,000 samples from 7 classes. Each sample is a 2,000-dimensional 1-d feature. The label of each ideal sample is also known. This dataset is not balanced where the trilinic class has as few as 14,000 samples while the cubic class has over 52,000 samples. Such data imbalance can be handled by the batch selection strategy in the training for the prototypical classifier. The imperfect dataset has only 1,494 samples from 7 classes, which is much fewer than the ideal dataset. 5-fold cross-validation is used for training the classifier and refiner. Furthermore, the theoretically computed ground-truth XRDs from materials scientist are also provided and are only used for evaluation purpose. For training the refiner, we have two settings where the class labels may or may not be known.

\noindent \textbf{Implementation details:} The refiner network, $\mathcal R_\phi$ is U-Net \cite{ronneberger2015u}. Since the inputs are 1-d XRD patterns of length 2000, all the 2-d layers in the U-Net are changed to 1-d layer while other configurations remain the same. The input $2000\times 1$ feature is convolved with $3\times 1$ filters that output 32 feature maps. The output is then passed through an encoder-decoder network structure with 4 convolutional and 4 deconvolutional layers. The output of the last layer passes through a $1\times 1$ convolutional layer which produces 1 feature map of size $2000\times 1$.

For the classifier $\mathcal C$, we use two structures DenseNet \cite{huang2017densely} and VGG \cite{simonyan2014very} to show that the imitation refinement framework works for different classifiers. For DenseNet, DenseNet-121 structuere is adopted, with 4 blocks where the numbers of dense-layers in each block are 6, 12, 24, 16. Each dense-layer is a composition of two BatchNorm-ReLU-Conv layers where the filter size is $5\times 1$. Growth rate is set to be 32. Note that the input pattern would go through Conv-BatchNorm-ReLU-Pooling layers first to produce a feature map that can be fed into subsequent blocks. VGG-19 is another classifier used in our experiments. We keep most configurations from the original paper except that all the 2-d layers are adapted to 1-d layers. The kernel size for the convolutional layers is $3\times 1$ and the kernel size for the max pooling layers is $10\times 1$. These changes are only made to fit the dimensionality of the input XRDs. In this application, function $\mathcal G$ is the last softmax layer of the classifier and $\mathcal F$ is the rest part of the classifier outputing the embeddings. We train the prototypical classifier for 100 epochs with batch size 512 (73 from each class and 1 more cubic to get sum 512). Function $\Psi(\cdot)$ is identity function and $\ell_{\mbox{reg}}$ uses $\ell_1$ norm. We use Adam to train the classifier with learning rate 0.001.

\begin{table}
\centering
\begin{tabular}{|c|c|c|}
\hline
Models & Standard & Prototypical \\
\hline
VGG-19 & 68.54\% & 69.01\%  \\
\hline
DenseNet & 67.74\% & 70.82\% \\
\hline
%{\em ip} & 0.1835 & 0.1872 & 0.1932 & \textbf{0.2306}\\
%\hline
\end{tabular}
%\vspace{-0.3em}
\caption{XRD patterns: A classifier pre-trained on the ideal dataset is tested on the imperfect data. The accuracies from standard and prototypical classifiers are given in the table. Prototypical classifiers perform better that the standard classifiers. These results are also used  as baselines.}
\label{pre_cls}
\end{table}

We first show the advantage of the prototypical classifier over the standard classifier with regard to the generalization by directly feeding the imperfect data into the classifiers pre-trained on the ideal dataset. Table \ref{pre_cls} shows the results. These results also serve as the baselines.

\begin{table}
\centering
\begin{tabular}{|c|c|}
\hline
Models & Accuracy \\
\hline
DWT & 71.28\% \\
\hline
ADDA & 73.78\% \\
\hline
GTA & 73.18\% \\
\hline
UNet+VGG & 71.37\% \\
\hline
UNet+proto-VGG & 73.98\% \\
\hline
UNet+DenseNet & 76.50\% \\
\hline
UNet+proto-DenseNet & \textbf{80.05\% }\\
\hline
\hline
non-targeted UNet+proto-VGG & 71.63\% \\
\hline
non-targeted UNet+proto-DenseNet & 74.74\% \\
\hline
%{\em ip} & 0.1835 & 0.1872 & 0.1932 & \textbf{0.2306}\\
%\hline
\end{tabular}
%\vspace{-0.3em}
\caption{Different accuracies from different methods or different settings. We also give the results in non-targeted cases. Our method outperforms the previous methods.}
\label{rfd_cls}
\vspace{-1em}
\end{table}

To show the improvement of the refined inputs with regard to the classification performance, Table \ref{rfd_cls} presents the label prediction accuracies from different methods or settings. Discrete wavelet transform \cite{cai1998different} is a widely used signal denoising technique in materials science domain. It removes high frequency noise and produces cleaner data. ADDA \cite{tzeng2017adversarial} is a recently proposed adversarial domain adaptation method aiming at learning different feature extraction networks for two similar domains. The embedding space learned by the two networks are aligned and can be used to predict labels. GTA \cite{sankaranarayanan2018generate} also learns a common embedding space that can be used for class prediction, as well as for training a GAN \cite{goodfellow2014generative} where the generator acts as a decoder decoding the embedding to a pattern in the raw feature space. The decoded pattern is then fed into a multi-class discriminator. As shown in the table, our methods outperform these state-of-art methods in this application where the imperfect data is not abundant. This table also shows that the refinement using the prototypical classifiers can achieve better results than using standard classifiers. The strength of our method when the labels are not available (non-targeted case) is demonstrated in the table as well. Compared to the two baselines ($69.01\%$ and $70.82\%$), the unsupervised training under the proposed framework gives promising results (71.63\% and 74.74\%). 

\noindent \textbf{Ablation study:} To show the combination of the different losses is necessary and meaningful, we show the  results in Table \ref{abl_cls} when either $\ell_{\mbox{proto}}$ or $\ell_{\mbox{pred}}$ is ablated. Note that all the reported numbers are averaged over 5 independent runs.

\begin{table}
\centering
\begin{tabular}{|c|c|c|c|c|}
\hline
Models & $\ell_{\mbox{proto}}$ & $\ell_{\mbox{pred}}$ & $\ell_{\mbox{reg}}$ & Accuracy \\
\hline
UNet+DenseNet & Y & N & Y & 79.33\% \\
\hline
UNet+DenseNet & N & Y & Y & 75.76\% \\
\hline
UNet+DenseNet & Y & Y & Y & \textbf{80.05\%} \\
\hline
%{\em ip} & 0.1835 & 0.1872 & 0.1932 & \textbf{0.2306}\\
%\hline
\end{tabular}
%\vspace{-0.3em}
\caption{The combination of 3 losses gives the best accuracy. ``Y'' stands for yes and ``N'' stands for no.}
\label{abl_cls}
\vspace{-1em}
\end{table}

\begin{figure}[!tbp]
  \centering
  \subfloat[GTA embedding space]{\includegraphics[width=0.25\textwidth]{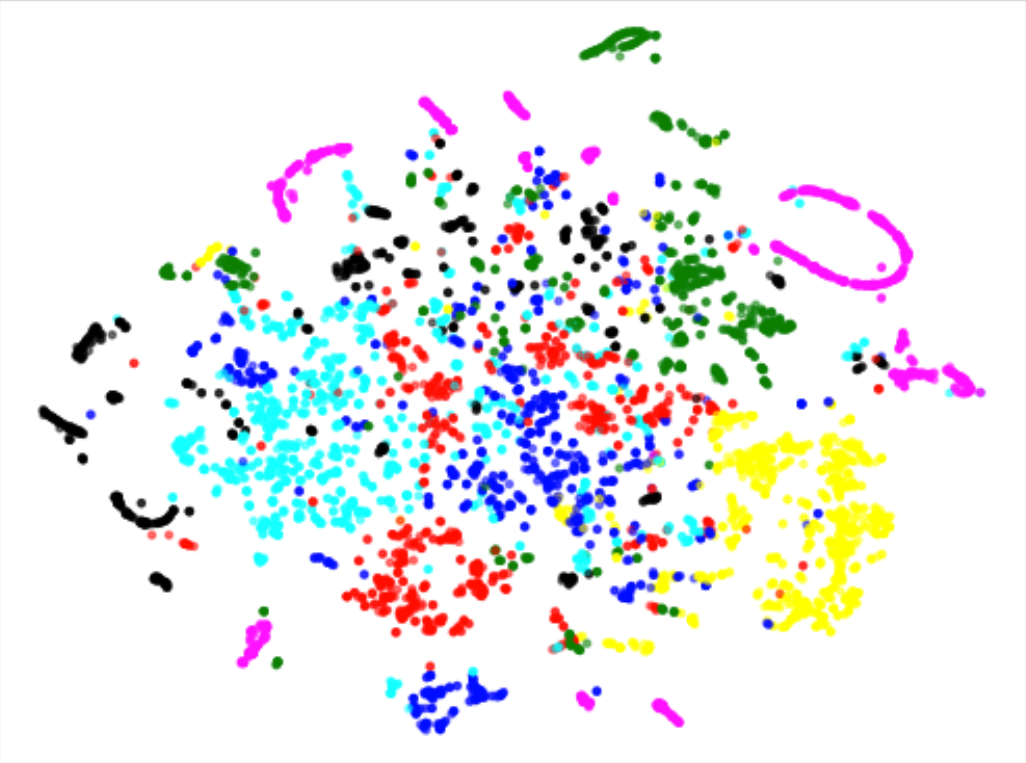}\label{fig:f1}}
  \subfloat[Our embedding space]{\includegraphics[width=0.25\textwidth]{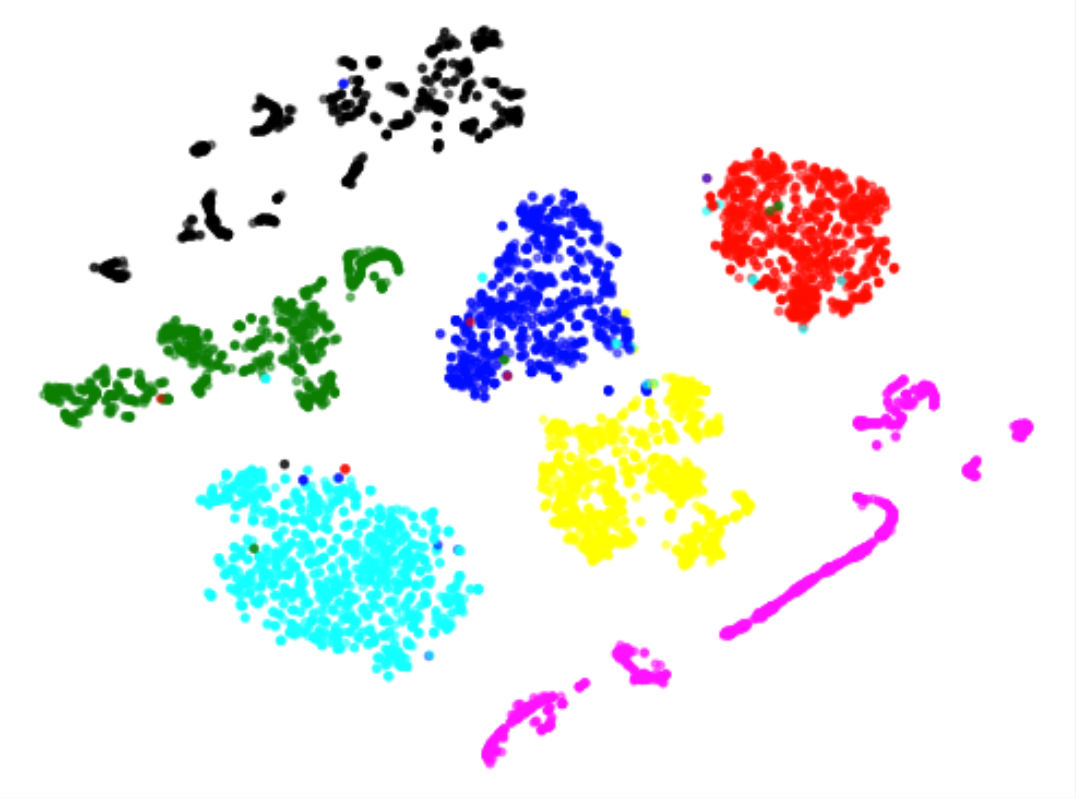}\label{fig:f2}}

  \caption{TSNE visualization of the embedding spaces learned by GTA and our imitation refinement. Each color represents a class. The embedding space from our method clusters the data better. }
  \label{emb_sp}
  \vspace{-1em}
\end{figure}

\noindent \textbf{Quantitative results} With respect to the quality of refined XRDs: We directly measure the differences between the refined XRDs and the ground-truth theoretical XRDs on 4 metrics, $\ell_1, \ell_2$, kl-divergence and cross correlation. We compute both the averages and the medians of the differences over all the test data (Table \ref{diff}). ADDA cannot produce refined XRDs, so it is skipped in this part of experiment. GTA does not perform well in the quality measurement since it does not consider the self-regularization loss or a prototypical loss and it actually learns a different refinement space. Besides, the embedding space learned by GTA is not as well-formed as imitation refinement (see Fig. \ref{emb_sp}).

\begin{figure}[t]
	\centering
	\includegraphics[width=0.48\textwidth]{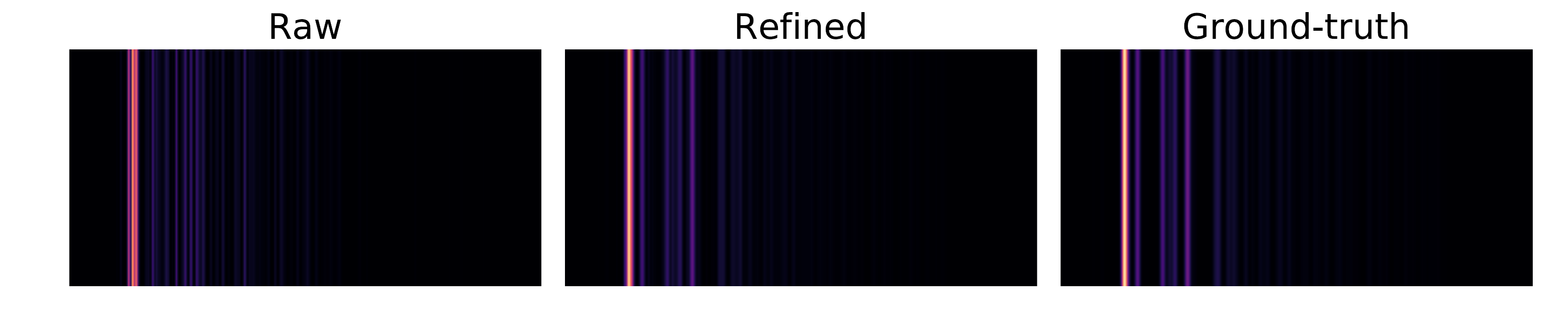}
    \includegraphics[width=0.48\textwidth]{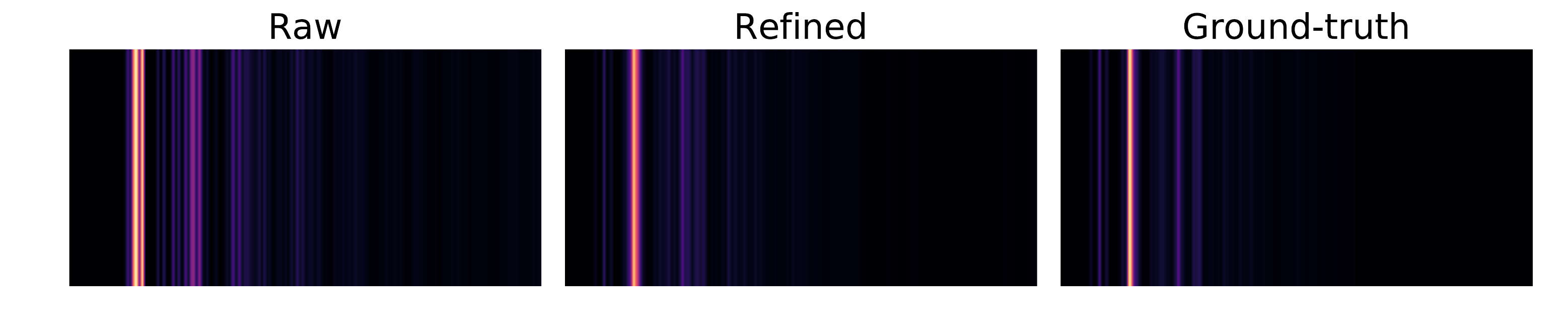}
    \caption{The visualization of the raw, refined and ground-truth XRDs for NbGa$_3$ and Mn$_4$Al$_{11}$ (best in color). The first one originally has 3 dim high peaks instead of 1 bright peak as in the ground-truth. Our method successfully squeezes the 3 peaks into one brightest peak. The second one originally does not have 3 peaks on the left to the brightest peak as ground-truth. Imitation refinement adds them.}
	\label{XRD}
    \vspace{-1em}
\end{figure}

\noindent \textbf{Qualitative analysis}: Fig. \ref{emb_sp} gives a comparison between the embedding space learned by GTA and imitation refinement. Fig. \ref{XRD} gives an example of the raw XRD, refined XRD and the ground-truth XRD for materials NbGa$_3$ and Mn$_4$Al$_{11}$.

\begin{table*}[htbp]
    \centering
    \begin{tabular}{|c|c|c|c|c|c|c|c|c|}
        \hline 
        & \multicolumn{4}{c|}{Average} & \multicolumn{4}{c|}{Median}\\
        \hline
        Models & $\ell_1$ & $\ell_2$ & $KL$ & NCC & $\ell_1$ & $\ell_2$ & $KL$ & NCC \\
        \hline
        Raw XRD & 38.993 & 4.913 & 41.027 & 24.415 & 26.749 & 1.718 & 12.195 & 20.870 \\
        \hline
        DWT & 37.459 & 4.722 & 37.920 & 24.305 & 25.642 & 1.706 & 11.972 & 20.884 \\
        \hline
        GTA & 89.265 & 9.498 & 48.263 & 20.991 & 90.914 & 6.861 & 23.306 & 20.618 \\
        \hline
        UNet+proto-VGG & 37.709 & 4.822 & 35.750 & 24.514 & 26.200 & \textbf{1.603} & \textbf{11.686} & 20.956 \\
        \hline
        non-targeted UNet+proto-VGG & 38.843 & 4.983 & 38.494 & 24.205 & 26.034 & 1.630 & 12.545 & 21.232 \\
        \hline
        UNet+proto-DenseNet & \textbf{36.744} & \textbf{4.382} & \textbf{31.767} & \textbf{25.827} & \textbf{25.101} & 1.671 & 11.945 & \textbf{22.481} \\
        \hline
        non-targeted UNet+proto-DenseNet & 37.119 & 4.797 & 36.246 & 24.364 & 27.235 & 1.754 & 11.834 & 21.079 \\
        \hline
    \end{tabular}
    \caption{Differences between the refined XRDs and the ground-truth XRDs on metrics $\ell_1, \ell_2, KL$ and normalized cross correlation (NCC). The difference between the raw XRDs and the ground-truth XRDs gives the baseline. The differences shown in the table are the averages or medians over all the test data. For $\ell_1, \ell_2, kl$, the smaller the better. For NCC, the larger, the better. The best results for each measure is in \textbf{bold}.}
    \label{diff}
    \vspace{-1em}
\end{table*}

\subsection{Hand-Written Digits}
%We first show the effectiveness of the proposed imitation refinement on the popular MNIST dataset \cite{mnist}, where a refiner is trained to correct ambiguous poorly written digits selected from the original dataset.

We also show the generality of our model in a hand-written digit refinement task. In this experiment, we show that if the ideal digits typeset in different fonts are given, imitation refinement can take advantage of these ideal digits to refine handwritten digits, so as to improve prediction accuracy as well as the readability.

\noindent \textbf{Ideal Digit Datasets:}
%\paragraph{Idealized Digits Datasets} 
We generate an ideal dataset containing images of digits typeset in five different fonts respectively: $\mathtt{Bradly~Hand}$, $\mathtt{Brush~Script}$, $\mathtt{Hannotate}$, $\mathtt{Times}$, and $\mathtt{Typewriter}$. 
We augment images in each font by vertically and horizontally shifting at most two pixels, and also rotating in range $[-20^{\circ}, 20^{\circ}]$. 
Each dataset contains 10250 images of digits. We merge these five datasets to a dataset $\mathtt{Synz}$ containing these synthetic digits.
%We also show the results in a extreme case where only $\mathtt{Times}$ digits are provided.

\begin{figure}[h]
    \centering
    \includegraphics[width=0.48\textwidth]{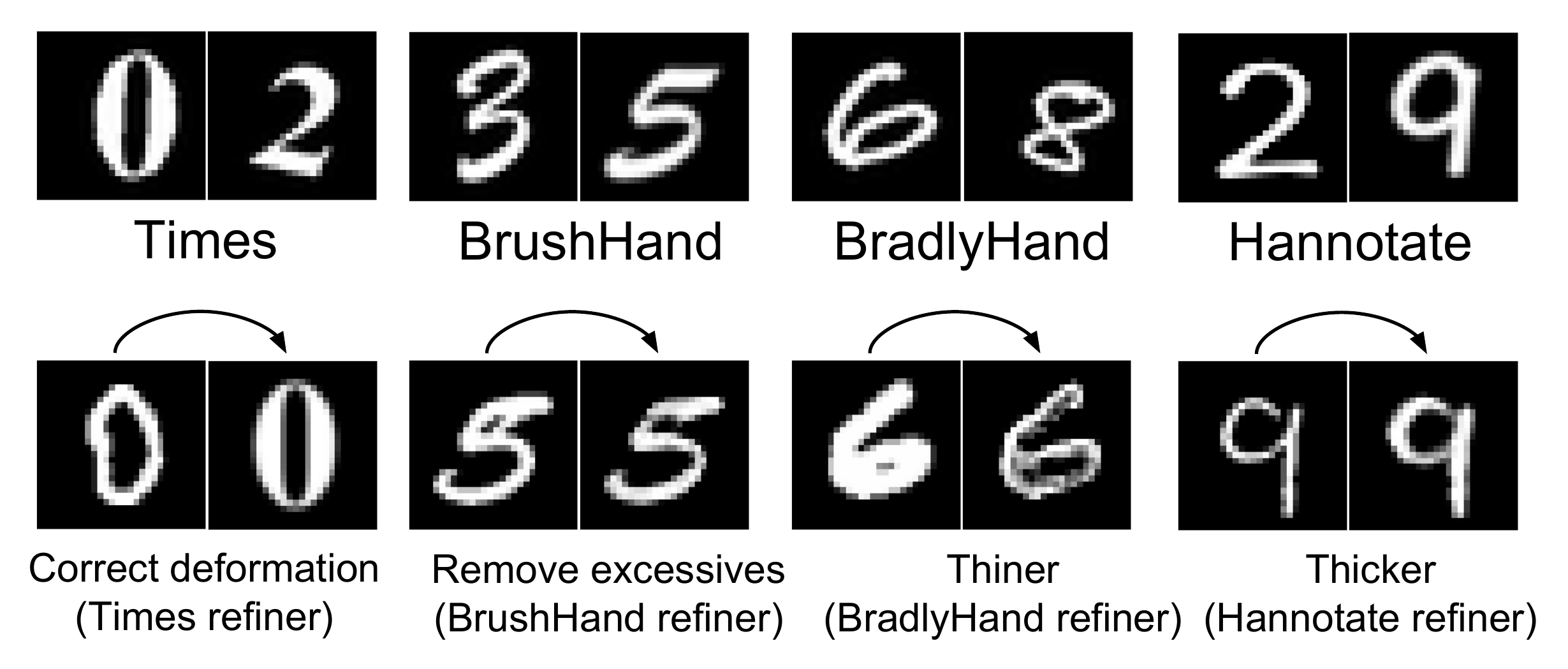}
    \caption{The first row displays example digits in the ideal dataset. The second row shows examples of imitatively refining MNIST handwritten digits to have the characteristics of different fonts.}
    \label{fig:fonts}
    \vspace{-1em}
\end{figure}

\noindent \textbf{Handwritten Digit Datasets:}
%\paragraph{Handwritten Digits Datasets} 
The MNIST dataset \cite{mnist} is used as our imperfect dataset, which contains 60,000 handwritten digits for the training and validation and another 10,000 digits for testing. 
%\newline
Some examples of digits in the \textit{ideal} dataset and imperfect dataset are shown in Fig. \ref{fig:fonts}. Our goal is to refine handwritten digits by mimicking the characteristics of computer generated fonts. We take 1 to 50 examples from each class to produce a small training set (10 to 500 data in total), which is then used for training. To allow reproducibility and reduce randomness in sparse data selection, we select first $N$ ($N\in \{1,2,5,10,20,30,40,50\}$) data in each class of the official MNIST training dataset. This dataset imitates data obtained under scenarios in which labeled data can be only sparsely obtained.

\begin{comment}
\begin{table}[htbp]
    \centering
    \begin{tabular}{|c|c|c|}
        \hline & $\mathtt{Times}$  &  $\mathtt{Synz}$\\ 
\hline        
Original accuracy &  25.03 & 71.21 \\
\hline Our method &  \bf{99.04} & 98.61 \\ 
\hline GTA &  89.59 & \bf{98.94} \\ 
\hline ADDA & 87.32 & 96.16 \\
\hline\hline Non-targeted &  36.82 &  73.71 \\
\hline
    \end{tabular}
    
    \caption{The improvement of identifiability of hand-written digits after imitation refinement. The first row shows the raw accuracies tested directly on MNIST dataset without any changes on the models or inputs. Row 2-4 show testing accuracy of our method and two baseline methods. Our method outperforms other models. In the extreme case of no available labels(non-targeted), imitation refinement can still improve the accuracies.
    }    
    \label{tab:mnist}
    \vspace{-1em}
\end{table}
\end{comment}

\noindent \textbf{Implementation details}
Most experimental settings remain the same as the previous materials discovery application except that the batch size in training the classifier is 100 and the optimizer is RMSprop with learning rate 0.01.

We first train the prototypical classifier $\mathcal C$ on the ideal digit datasets. $\mathcal C$ classifies instances in the ideal dataset into 10 categories (0-9). Similar to the earlier experiment, we take DenseNet-121 as the basic network structure. In order to fit the smaller resolution input of this experiment, we replaced the first downsampling convolution with a convolution of stride 1, and we remove the first max-pooling layer and the first dense block. The rest remains unchanged. 
This classifier achieves over $99.9\%$ test accuracy on the ideal datasets.

We use a U-Net structured model as the refiner on the MNIST dataset. Note this network is very small, with only 60k parameters (compared to 900k parameters in DenseNet). Therefore adding very little overhead to the existing classifier. When training, we consider two cases: a targeted case in which the labels of handwritten digits are given and a non-target case where the labels are not given. Note that in both cases, the ideal counterparts of the imperfect digits are unknown. By training our network end-to-end, the network is able to learn how to best perform this refinement. This will be shown quantitatively in the next section.

% The evaluation loss $\eta_{\mathcal C}$ in the targeted case is the negative log-likelihood, and in the non-targeted case it is the entropy of digit predictions.
% The discrepancy distance $\delta_{\mathcal C}(x, \mathcal R(x))$ in this experiment is defined by the $\ell_1$-distance between the corresponding features maps extracted from the second convolutional layer of the classifiers. 
% We set the $\lambda = 0.55$ in the objective function (\ref{eq:opt}) to balance the refinement simplicity and effectiveness. An Adam optimizer with initial learning rate $10^{-4}$ is used to train the refiners with batch size 50 for total ten epochs.

\subsubsection{Improvement of accuracy:}

\begin{figure}[htbp]
    \centering
    \includegraphics[width=0.95\linewidth]{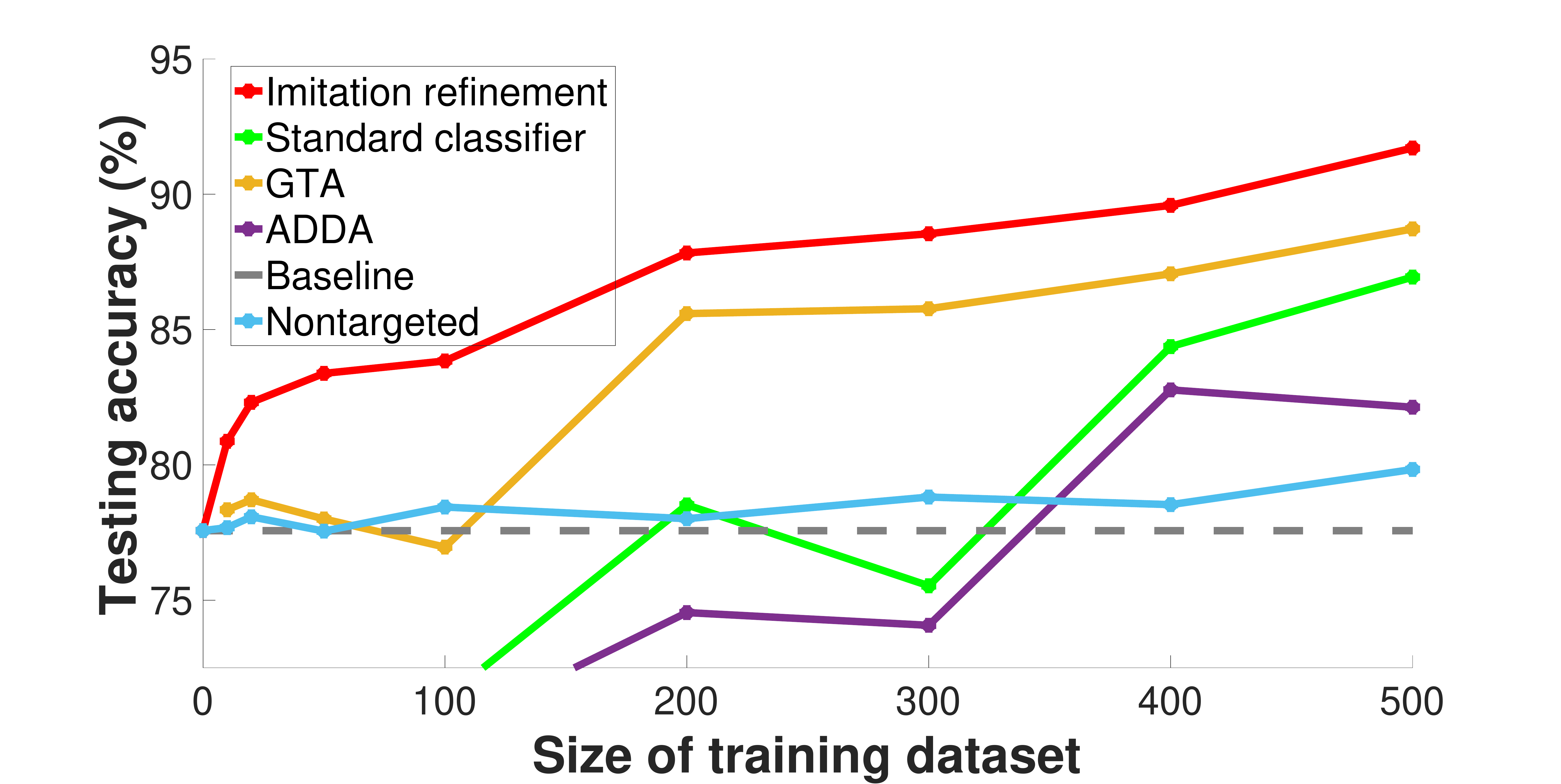}
    \caption{As the number of training examples from  MNIST dataset increases, the test accuracies of all the methods can increase gradually. However, only imitation refinement consistently outperforms the baseline and other methods.
%However, taking advantage of imitation refinement, the ideal classifier can better recognize handwritten digits when not many imperfect data are available. Note accuracy increases even with as little as 1 sample per class. 
    }
    \label{fig:num_exp}
    \vspace{-1em}
\end{figure}

As mentioned previously, we take 1 to 50 examples from each class to produce a small training set (0.02\%-1\% of entire MNIST training dataset). Fig. \ref{fig:num_exp} shows the improvement of identifiability of hand-written digits after imitation refinement. 

The baseline is given in the gray dash line. Directly feeding the mnist test data into the pre-trained prototypical classifier gives an accuracy of  77.57\%. This result is better than the accuracy obtained when using a standard classifier with the same structure, which is only  70.26\%. The green line is achieved by training a standard classifier from scratch on the very few given training data from scratch. When the number of samples from each class is smaller than 10, the classifier cannot learn anything. But as the number of training samples grows, the accuracy also increases. 
%We call this classifier the "real-world" classifier since it learns the knowledge from the "real-world" imperfect data without any prior knowledge. 
This classifier cannot outperform the baseline stably until after seeing over 40 samples from each class. The results given by another two compared methods, ADDA and GTA are also shown here. ADDA corresponds to the purple line and its performance is not very promising in this scarce data scenario. GTA performs better in that it learns a more meaningful embedding space which helps the classification even when the data amount is not rich. 
The results of imitation refinement are given by the red line.
% consistently outperforms other methods and the baseline. 
%Note that our method outperforms the baseline incrementally from the start.
Starting from the baseline accuracy, imitation refinement consistently outperforms the other methods and it improves with the increase of training dataset size.
%In addition, we also train imitation refinement in the non-targeted case where only 50 examples are given from each class. The test gives 
% We want to stress on that instead of modifying or fine-tuning classifiers to achieve better classification results, we only refine the inputs and obtain the promising results. 
Note that imitation refinement obtains better results by modifying and refining the classifier's input, instead of changing the classifier's model. 
This illustrates that the prior knowledge learned by the classifier can actually help produce meaningful refined patterns and is a key difference between our work and related methods such as transfer learning. Fig. \ref{fig:fonts} also shows the meaningful features learned in the refined digits. For instance, ``0" learns a meaningful deformation to mimic the $\mathtt{Times}$ style and ``1" learns the essential bottom bar in the refined digit while preserving most contents in the raw image. We also perform imitation refinement in the non-targeted case and it outperforms the baseline and increases the baseline accuracy with a margin of over 2\% when 50 samples are given per class.

%The best performance is achieved by the ideal classifier trained on the dataset $\mathtt{Synz}$, with 71.21\% accuracy. 
%After applying our refinement, classification accuracies are consistently over 98\%, meaning almost all refined handwritten digits can be recognized by all these ideal classifiers. This significant increase in accuracy, ranging from 27.4\% to 78.8\%, surpassed all our baselines.

%In the extreme case where no label is given, our method can increase accuracy with a small margin (Row 5). 

\begin{comment}
\subsubsection{The power of prior knowledge}
We mimic some real world scenarios in which labeled real data is hard to obtain. Specifically, we take 1 to 50 examples from each class to produce a very small training set (10 to 500 data in total, 0.02\%-1\% of entire MNIST training dataset), which is then used for training. This dataset imitates data obtained under scenarios where only a few labeled data can be obtained. We show that guided by knowledge gained from observing large amount of synthetic data, even such a small dataset can be effectively used to learn a high quality refiner. 

As a baseline, we train a real-world classifier without prior knowledge with the same structure directly on the same dataset. The results are shown in Fig. \ref{fig:num_exp}. Our method clearly performs better when using a limited dataset. The accuracy improves to close 80\% with only 20-50 training data, while the baseline lingers around 60\%. 
\end{comment}
\section{Related Work}

Imitation refinement is related to data denoising \cite{xie2012image} and data restoration \cite{dong2014learning}, which improve noisy or corrupted data. 
One key difference is that imitation refinement does not require imperfect data to  be paired with its cleaned version while training. 
%Also, it does not assume a noise model. In fact, the ground-truth of the data in our imperfect dataset is unknown during training. 
%In fact, members in our imperfect dataset are not related to those in the ideal dataset.
%% completely different. 
We directly train an imitation refiner on the imperfect dataset, for which we don't know the ground truth. 
Also, the ideal dataset does not provide ground truth counterparts for the imperfect data.
%re is no correspondence between members of the ideal datest of which no ground truth is known, 
%since 
%the members of the ideal dataset 
%and imperfect dataset refer to different entities. 
%% contain no ground truth counterparts of the imperfect data.
%
Our work is also related to style transfer \cite{gatys2016image}, 
%which also requires paired images prior and after the style transformation. 
which requires paired images or cycle consistency (a bijection between two domains) for the transformation.
%% However, imitation refinement does not only imitate the ``style" of ``ideal" data while keeping the ``content" of the imperfect data.
%% It imitates the ``ideal" data to improve the imperfect data to better fits the prior knowledge or physical rules. 
%
Besides, transfer learning \cite{pan2010survey} and domain adaptation \cite{glorot2011domain} modify models while our classifier is fixed.
Additionally, imitation refinement has notable differences with conventional inverse classification \cite{mannino2000cost,aggarwal2010inverse}, which uses inverted statistics to complete partial data or solve an optimization problem for each test sample respectively. 
Imitation refinement differs from these methods since  it incorporates the knowledge embedded in a pre-trained classifier into the  refiner, which can generalize to the unseen imperfect data. 
Further, imitation refinement allows for high-level data modification, while inverse classification often only changes data attributes.
%
%Finally, imitation refinement is different from GAN \cite{gan} since its goal is not to refine imperfect data such that it matches the ideal data. 
Finally, imitation refinement is different from GAN \cite{goodfellow2014generative} since its goal is not to refine imperfect data such that a
discriminator cannot differentiate them from ideal data. Instead,  imitation refinement    only applies small modifications to the imperfect data to  reflect the  fundamental characteristics  of the ideal data, captured by the classifier trained on them. Several recently proposed domain adaptation methods using GAN \cite{tzeng2017adversarial,sankaranarayanan2018generate} seek to find a common space for both target domain and source domain. Though they give nice classification results, the performance on the refinement of the raw inputs is not their focus. Besides, these methods typically require a certain amount of data. As shown in the experimental section, they do not perform very well when only a limited amount of imperfect training data are given.
Prototypical network \cite{snell2017prototypical} is closely related to our work. It learns a meaningful embedding space formed by the support sets from each batch. However, their work use the embedding space for classification while we mainly use it for refinement.

% However, the space might change for different batches while our prototypical classifier learns a stable embedding space for all the data.
\section{Conclusion and Future Work}

{\em Imitation refinement} improves the quality of imperfect data by imitating  {\em ideal data}. Using the prior knowledge captured by a {\em prototypical classifier} trained on an ideal dataset, a refiner learns to apply modifications to imperfect data to improve their qualities. A general end-to-end neural framework is proposed to address this refinement task and gives promising results in two applications: handwritten digits and XRD pattern refinement. Imitation refinement improves readability and accuracy of identifying handwritten digits and refines the XRDs to be closer to the ground-truth patterns. This work has a potential to save lots of manual work for material scientists. We also show that imitation refinement could work even if labels are not provided. Imitation refinement is easily adaptable to  other  different situations, such as crowd-sourcing tasks, where the raw data are often imperfect. The refiner and classifier are also replaceable components and we have shown that the imitation refinement framework can incorporate prior knowledge efficiently. We hope our work will stimulate additional imitation refinement efforts.

\bibliography{aaai19}
\bibliographystyle{aaai}

\end{document}